\documentclass{article}
\usepackage{spconf,amsmath,graphicx,hyperref}
\usepackage{amssymb}
\usepackage{booktabs}
\usepackage{hyperref}

\usepackage{amsfonts,dsfont}
\usepackage{algorithmic}
\usepackage{graphicx,color}
\usepackage{textcomp}
\usepackage{xcolor}
\usepackage{algorithm,algorithmic}
\usepackage{arydshln}
\usepackage{float} 
\usepackage{mathtools}
\usepackage{caption}
\usepackage{subcaption}
\usepackage{cleveref}
\usepackage{multirow}

\crefname{figure}{Fig.}{Figs.}
\crefname{table}{Tab.}{Tabs.}
\crefname{equation}{Eq.}{Eqs.}
\crefname{section}{Sec.}{Secs.}

\Crefname{figure}{Figure}{Figures}
\Crefname{table}{Table}{Tables}
\Crefname{equation}{Equation}{Equations}
\Crefname{section}{Section}{Sections}

\newcommand*{\Scale}[2][4]{\scalebox{#1}{$#2$}}%

\DeclareMathOperator{\Tr}{Tr}

\DeclareMathOperator{\sign}{sign}

\usepackage{xcolor}

\title{DISCRIMINANT LEARNING-BASED COLORSPACE FOR BLADE SEGMENTATION}
\name{Raül~Pérez-Gonzalo$^{1,2}$, Andreas~Espersen$^{2}$ and Antonio~Agudo$^{1}$\thanks{This work has been partially supported by the project GRAVATAR PID2023-151184OB-I00 funded by MCIU/AEI/10.13039/501100011033 and ERDF, UE; and by GreenVAR project of the Fundación Ramón Areces. }}
\address{$^{1}$Institut de Robòtica i Informàtica Industrial, CSIC-UPC, Barcelona, Spain\\$^{2}$Wind Power LAB, Copenhagen, Denmark}
\begin{document}
%
\maketitle
\begin{abstract}
Suboptimal color representation often hinders accurate image segmentation, yet many modern algorithms neglect this critical preprocessing step. This work presents a novel multidimensional nonlinear discriminant analysis algorithm, Colorspace Discriminant Analysis (CSDA), for improved segmentation. Extending Linear Discriminant Analysis into a deep learning context, CSDA customizes color representation by maximizing multidimensional signed inter-class separability while minimizing intra-class variability through a generalized discriminative loss. To ensure stable training, we introduce three alternative losses that enable end-to-end optimization of both the discriminative colorspace and segmentation process. Experiments on wind turbine blade data demonstrate significant accuracy gains, emphasizing the importance of tailored preprocessing in domain-specific segmentation. 
\end{abstract}

\section{Introduction}
\vspace{-0.15cm}

Image segmentation partitions an image into homogeneous regions and represents a fundamental problem in image processing. Among visual features, color is crucial in achieving comprehensive image representation and reliable segmentation~\cite{survey_color,importance_cs}. Consequently, selecting an appropriate colorspace is vital to achieve high segmentation performance.

Over the past decade, deep-learning techniques have driven significant progress in image segmentation~\cite{deeplabv3+,segment2025-plosone}, including domain-specific industry solutions~\cite{bunet}. Most approaches rely on encoder–decoder architectures~\cite{sw,unetformer} and attention mechanisms for enhanced performance~\cite{mask2former,resnest}. Recent models target efficiency in resource-constrained settings~\cite{mobilevit,efficientformer}, while zero-shot methods such as SAM~\cite{sam} reflect the versatility of transformer-based segmentation~\cite{clipseg,diffseg}.

Despite these advances, segmentation algorithms rarely optimize the underlying colorspace for domain-specific needs. Traditional spaces such as RGB are designed for general-purpose imaging~\cite{color_seg_survey} and may fail to capture subtle cues critical in specialized contexts. Moreover, hybrid spaces that combine components of standard colorspaces~\cite{discriminative_channel_cs,color_seg_survey,importance_cs_classif} also lack class-specific discriminability.

\begin{figure}[t!]
\centering
\includegraphics[width=\linewidth]{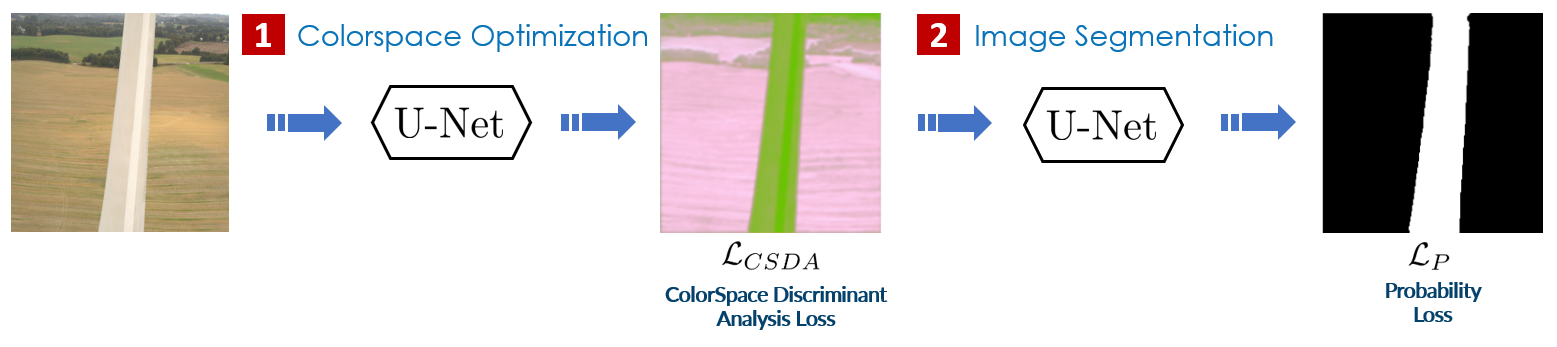} 
\vspace{-0.85cm}
\caption{\textbf{CSDA framework.} Joint optimization of colorspace and segmentation via discriminant and probability losses.}
\label{fig:colorspace_model}
\vspace{-0.5cm}
\end{figure}

This paper addresses this gap by introducing a domain-specific colorspace tailored for the target application. Without loss of generality, we focus on wind turbine blade segmentation, where accurately isolating blades from the background is key for downstream defect detection and assessment tasks~\cite{PerezGonzaloIcip2024,3d-wind,perezGonzaloWACV26}. This focus is motivated by typical failure cases, where shadowed blade regions appear with intensities opposite to those illuminated, leading to segmentation errors~\cite{bunet}. Inspired by prior work in face recognition that uses class-based color information for discriminant analysis~\cite{discriminant_cs_baseline1,discriminant_cs_tensor,discriminant_cs_luminance}, we build on Linear Discriminant Analysis (LDA)~\cite{lda} to develop ColorSpace Discriminant Analysis (CSDA). CSDA is a learnable nonlinear transformation that integrates discriminant colorspace learning into an end-to-end segmentation framework, outlined in \cref{fig:colorspace_model}. A neural network jointly optimizes colorspace representation and segmentation, enhancing class separability in the learned embedding.



Compared with existing face recognition methods, CSDA proposes a multidimensional, nonlinear setting via deep discriminant analysis (DDA)~\cite{fisher_simplified,relu_lda}. Our formulation introduces a signed between-class variance matrix and avoids matrix inversion through three alternative CSDA losses, improving training stability and performance. Experimental evidence demonstrates that CSDA offers a robust and interpretable solution for colorspace optimization, achieving superior performance over current state-of-the-art baselines.

\vspace{-0.25cm}
\section{Colorspace Discriminant Analysis} \label{sec:methods}
\vspace{-0.15cm}

We propose a model that transforms input images into a new colorspace where blade and non-blade classes are maximally separable. Our approach builds on LDA~\cite{lda}, which seeks a linear projection that maximizes class separability based on the Fisher criterion: the ratio of between-class $s_b$ to within-class variance $s_w$. In the two-class case, $C_i$ with $i\in\{0,1\}$, this is defined as:

\vspace{-0.4cm}
\begin{align} \label{eq:fisher}
\max \frac{s_b}{s_w} = \frac{(\mu_{C_0} - \mu_{C_1})^2}{s^2_{C_0}+s^2_{C_1}}~,
\end{align}
where $\mu_{C_i}$ and $s^2_{C_i}$ denote the class mean and variance, respectively. We extend this principle to a multidimensional setting and further enhance discriminative power through nonlinear transformations optimized via gradient descent. The resulting CSDA framework integrates these ideas into an end-to-end model.

\vspace{-0.15cm}
\subsection{Multidimensional Discriminant Analysis} \label{sec:fisher_multi}
\vspace{-0.15cm}

LDA transforms the original feature space into a lower-dimensional one that emphasizes class separability, facilitating the classification of the classes. The dimensionality of the projected space is at most $C-1$, where $C$ is the number of classes. In a binary classification, this results in a single projection line. However, this reduction in dimensionality may discard useful discriminative information.

To address this, we propose an extended multidimensional LDA criterion that maximizes the trace of the between-class scatter $\mathbf{S}_b$ relative to the within-class scatter $\mathbf{S}_w$:

\vspace{-0.35cm}
\begin{align} \label{eq:fisher_multi}
\max \Tr \left( \mathbf{S}_w^{-1} \mathbf{S}_b \right) \text{ with }
 \begin{cases} 
\Scale[0.85]{\mathbf{S}_b} = \Scale[0.85]{(\boldsymbol{\mu}_{C_0} - \boldsymbol{\mu}_{C_1})(\boldsymbol{\mu}_{C_0} - \boldsymbol{\mu}_{C_1})^{\top}}     \\ 
  \Scale[0.85]{\mathbf{S}_w} = \Scale[0.85]{\mathbf{S}^2_{C_0}+\mathbf{S}^2_{C_1}}   \nonumber 
\end{cases}~,
\end{align} 
where $\boldsymbol{\mu}_{C_i}$ and $\mathbf{S}^2_{C_i}$ denote the mean vector and covariance matrix of class $C_i$, respectively. This criterion seeks to enhance class separability in a multidimensional space, retaining discriminative information for classification.  

\vspace{-0.15cm}
\subsection{Extending LDA to a Nonlinear Setting} \label{sec:gradient}
\vspace{-0.15cm}

The multidimensional Fisher criterion from \cref{eq:fisher_multi} is invariant to any nonsingular linear transformation $\mathbf{W}$. Thus, if classes are not separable in the original space, linear mappings cannot improve class separability, due to the trace remains unchanged under such transformations:


\vspace{-0.3cm}
\begin{align} 
& \Scale[0.8]{\Tr \left( (\mathbf{W}^{\top}\mathbf{S}_w\mathbf{W})^{-1} (\mathbf{W}^{\top}\mathbf{S}_b\mathbf{W}) \right)} = \Scale[0.8]{\Tr \left( (\mathbf{W}^{-1}\mathbf{S}_w^{-1}(\mathbf{W}^{\top})^{-1} \mathbf{W}^{\top}\mathbf{S}_b\mathbf{W}  \right)} \nonumber \\
&= \Scale[0.8]{\Tr \left( \mathbf{W}^{-1}\mathbf{S}_w^{-1}\mathbf{S}_b\mathbf{W} \right)} = \Scale[0.8]{\Tr \left( \mathbf{W}\mathbf{W}^{-1}\mathbf{S}_w^{-1}\mathbf{S}_b \right)} = \Scale[0.8]{\Tr \left( \mathbf{S}_w^{-1} \mathbf{S}_b \right)}~.
\end{align}

This motivates the need for nonlinear approaches, which can enhance discriminative power while preserving multidimensionality. In this setting, analytical optimization is intractable, so we resort to gradient-based methods. However, directly optimizing the Fisher criterion (\cref{eq:fisher_multi}) can lead to instability: the gradient of the between-class variance $\nabla \mathbf{S}b$ may alternate in sign across training steps, as it depends on the class means' relative positions. Letting $\mathbf{u} := (\boldsymbol{\mu}{C_0} - \boldsymbol{\mu}_{C_1})$, the sign of $\nabla \mathbf{S}_b$ is then given by:

\vspace{-0.5cm}
\begin{align}
\Scale[0.83]{
 \nabla(\mathbf{u}\mathbf{u}^{\top}) = (\nabla \mathbf{u})\mathbf{u}^{\top} +\mathbf{u}(\nabla\mathbf{u}^{\top})
 \begin{cases} 
< 0 \text{ if } \mathbf{u} < 0 \text{; } \boldsymbol{\mu}_{C_1} > \boldsymbol{\mu}_{C_0} ~\\ 
> 0 \text{ if } \mathbf{u} > 0 \text{; }  \boldsymbol{\mu}_{C_0} > \boldsymbol{\mu}_{C_1} ~
\end{cases}~.
}
\end{align}

To address inconsistent training updates, we adopt a signed between-class variance to ensure consistent gradient directionality. Let $D^{\sign}(\mathbf{u})$ be a matrix whose diagonal captures the sign of each component of $\mathbf{u}$, denoted by $\sign(u_j)$, and whose off-diagonal elements are one, then we obtain an adapted discriminant function to improve class separability in a nonlinear, multidimensional setting:

\vspace{-0.4cm}
\begin{equation} \label{eq:fisher_multi_sign}
\max \Tr \left( \mathbf{S}_w^{-1} D^{\sign}(\boldsymbol{\mu}_{C_0} - \boldsymbol{\mu}_{C_1}) \odot \mathbf{S}_b \right), \,\,\text{with}
\end{equation}

\vspace{-0.65cm}
\begin{align} 
\Scale[0.95]{D^{\sign}(\mathbf{u}) = \begin{pmatrix}
\sign(u_1) & 1 & \cdots & 1 \\
1 & \sign(u_2) & \cdots & 1 \\
\vdots & \vdots & \ddots & \vdots \\
1 & 1 & \cdots & \sign(u_{d_{CS}})
\end{pmatrix} } ~, \nonumber
\end{align}
where $d_{CS}$ is the dimensionality of the projected colorspace. 

The signed between-class variance is computed via the Hadamard product $\odot$ of $D^{\sign}(\mathbf{u})$ and $\mathbf{S}_b$. This new criterion introduces directionality by encoding class order information into the loss, ensuring a consistent mapping in which, for each  feature dimension $j \in \{1,\ldots,d_{CS}\}$, samples from $C_0$ are mapped to higher values than those from $C_1$, i.e., $y_{0j} > y_{1j}$.

\vspace{-0.15cm}
\subsection{Colorspace Discriminant Analysis Losses} \label{sec:losses}
\vspace{-0.15cm}

To optimize the colorspace, we propose three alternative loss functions derived from \cref{eq:fisher_multi_sign} that avoid the instability introduced by matrix inversion. These losses allow direct optimization of the colorspace for improved class separability in a nonlinear deep network setting, ensuring that the learned feature space maximizes discriminative power across classes.

{\bf CSDA Loss}. The first loss adapts \cref{eq:fisher_multi_sign} into a minimization form by retaining both between- and within-class variance in the numerator, avoiding matrix inversion. The signed between-class variance enforces an ordering in which positive-class components are consistently mapped above negative-class components, i.e., $y_{1j} > y_{0j}$:


\vspace{-0.4cm}
\begin{align} \label{eq:dda_loss}
\mathcal L_{CSDA} = \Tr \left( \mathbf{S}_w D^{\sign}(\boldsymbol{\mu}_{C_0} - \boldsymbol{\mu}_{C_1}) \odot \mathbf{S}_b \right) ~.
\end{align}

{\bf Logarithmic CSDA Loss}. An analogous procedure is applied to the separability measure $\frac{\Tr(\mathbf{S}b)}{\Tr(\mathbf{S}w)}$~\cite{fukunaga}. We adopt the signed between-class variance and convert this ratio into an additive form via logarithms. Since this signed matrix has $d_{CS}$ diagonal terms within $[-1,1]$, we shift the argument by $d_{CS} + \epsilon$ to keep the logarithm well defined. Finally, a scalar $\lambda_F$ controls the trade-off between both variance terms as:

\vspace{-0.4cm}
\begin{align} \label{eq:dda_loss_ln}
 \mathcal L_{CSDA}^{(\ln)} &= \ln\left( d_{CS} + \epsilon + \Tr \left( D^{\sign}(\boldsymbol{\mu}_{C_0} - \boldsymbol{\mu}_{C_1}) \odot   \mathbf{S}_b\right) \right) \nonumber\\ 
 & \hspace{0.35cm} + \lambda_F \ln\left( d_{CS} + \epsilon + \Tr \left(  \mathbf{S}_w \right)\right)  .
\end{align}

{\bf Delta CSDA Loss}. Instead of minimizing a ratio, another effective separability criterion is to optimize the sum of between-class separation and within-class compactness $\Tr(\mathbf{S}_b)+\Tr(\mathbf{S}_w)$~\cite{fukunaga}, providing a simple but effective loss:
\begin{align} \label{eq:dda_loss_delta}
\mathcal L_{CSDA}^{(\Delta)} = \Tr \left( D^{\sign}(\boldsymbol{\mu}_{C_0} - \boldsymbol{\mu}_{C_1}) \odot \mathbf{S}_b\right) + \lambda_F \Tr \left(  \mathbf{S}_w \right) ~.
\end{align}
\vspace{-0.99cm}

\subsection{Model Design} \label{sec:model}
\vspace{-0.1cm}

Both colorspace and segmentation models use a U-Net~\cite{unet} and are trained end-to-end (see \cref{fig:colorspace_model}). The colorspace model transforms an RGB image into an optimized representation $\mathbf{Y}$ that serves as input to the segmentation branch, which predicts the mask $\hat{\mathbf{M}}$ given the ground truth $\mathbf{M}$.

The total loss $\mathcal L_{T} = \mathcal L_{P} + \lambda_P \mathcal L_{DDA}$ combines our proposed discriminant losses $\mathcal{L}_{DDA}$ from \cref{sec:losses} with a probabilistic segmentation loss $\mathcal{L}_P$, balanced by $\lambda_P$. Specifically, we adopt the focal loss~\cite{focal} for $\mathcal{L}_P$, which emphasizes low-confidence predictions through exponential weighting.


\vspace{-0.25cm}
\subsection{Implementation Details} \label{sec:implementation}
\vspace{-0.05cm}

Input images are resized to $256\times256$, min-max normalized, and augmented with flipping and cropping. We use Adam optimizer with a learning rate of $10^{-4}$ and batch size of $8$. Our custom scheduler decreases the learning rate after three epochs without validation improvement. Pixel probabilities $\hat m$ are generated using the sigmoid function. We set $\epsilon=10^{-8}$ in \cref{eq:dda_loss_ln} for stable training. Balancing factors are $\lambda_P=1.3$ for focal and CSDA losses, and $\lambda_F=0.5$ for variance terms. 



    



\begin{figure}[t!]
    \centering
    \resizebox{1.01\linewidth}{!}{
    \hspace{-0.35cm} \begin{tabular}{@{}c@{\hspace{0.05cm}}|@{}c@{}}
    \begin{tabular}{@{}c@{}}
            \vspace{0.25cm} \text{}
            \includegraphics[width=0.365\linewidth]{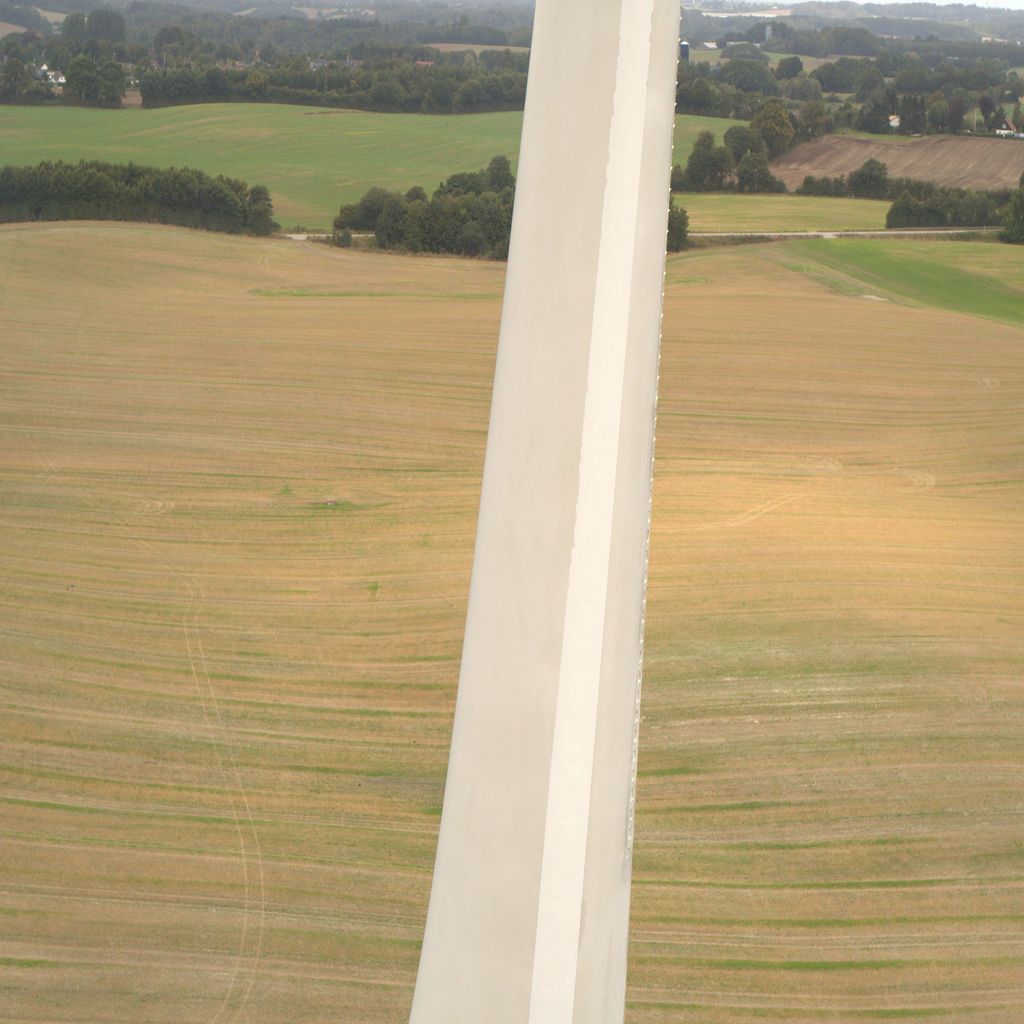}
         \\
            \hspace{0.14cm}  \includegraphics[width=0.365\linewidth] {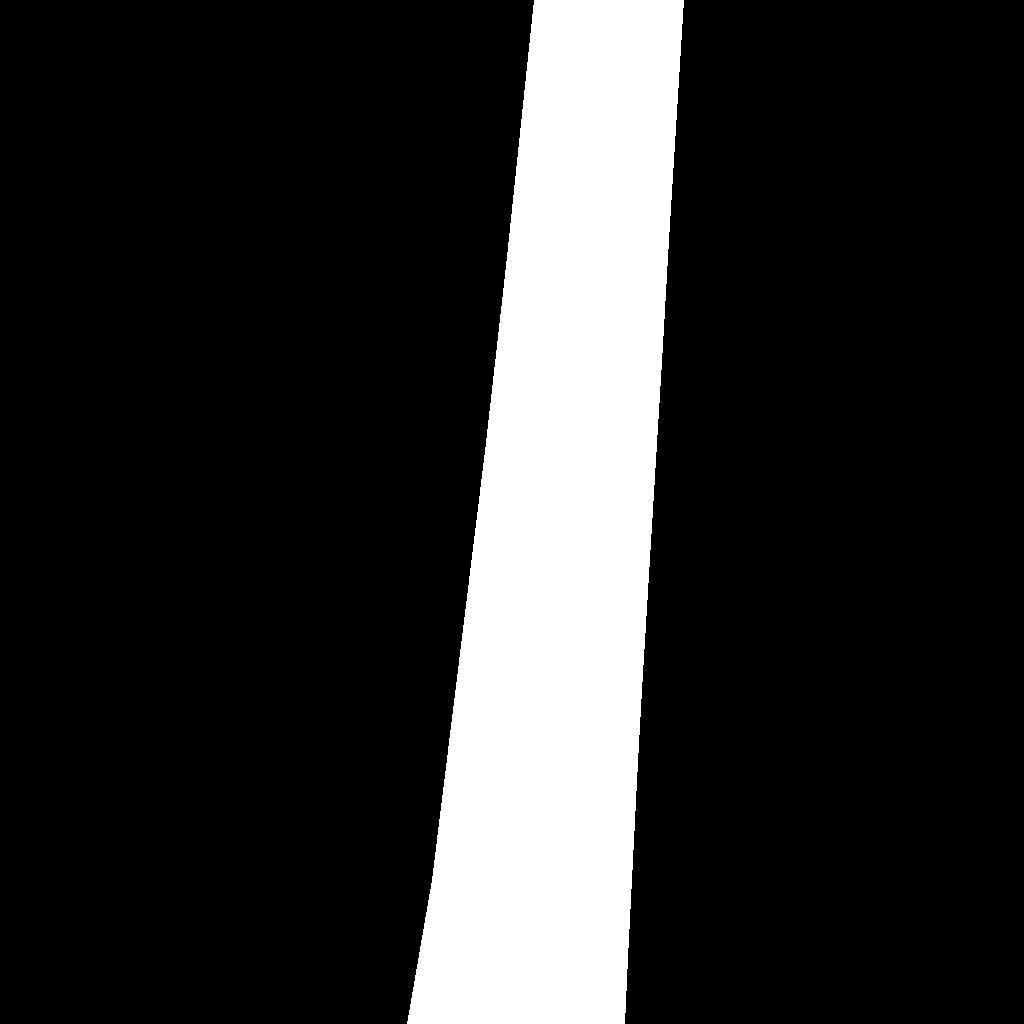} 
            \vspace{-0.25cm} \text{}
    \end{tabular} &
    
    \begin{tabular}{@{\hspace{0.06cm}}c@{\hspace{0.15cm}}:c@{}}
        \rotatebox{90}{\hspace{.38cm}\normalsize $d_{CS} = 1$}
        \begin{subfigure}[b]{0.471\linewidth}
            \centering
            \includegraphics[width=\linewidth]{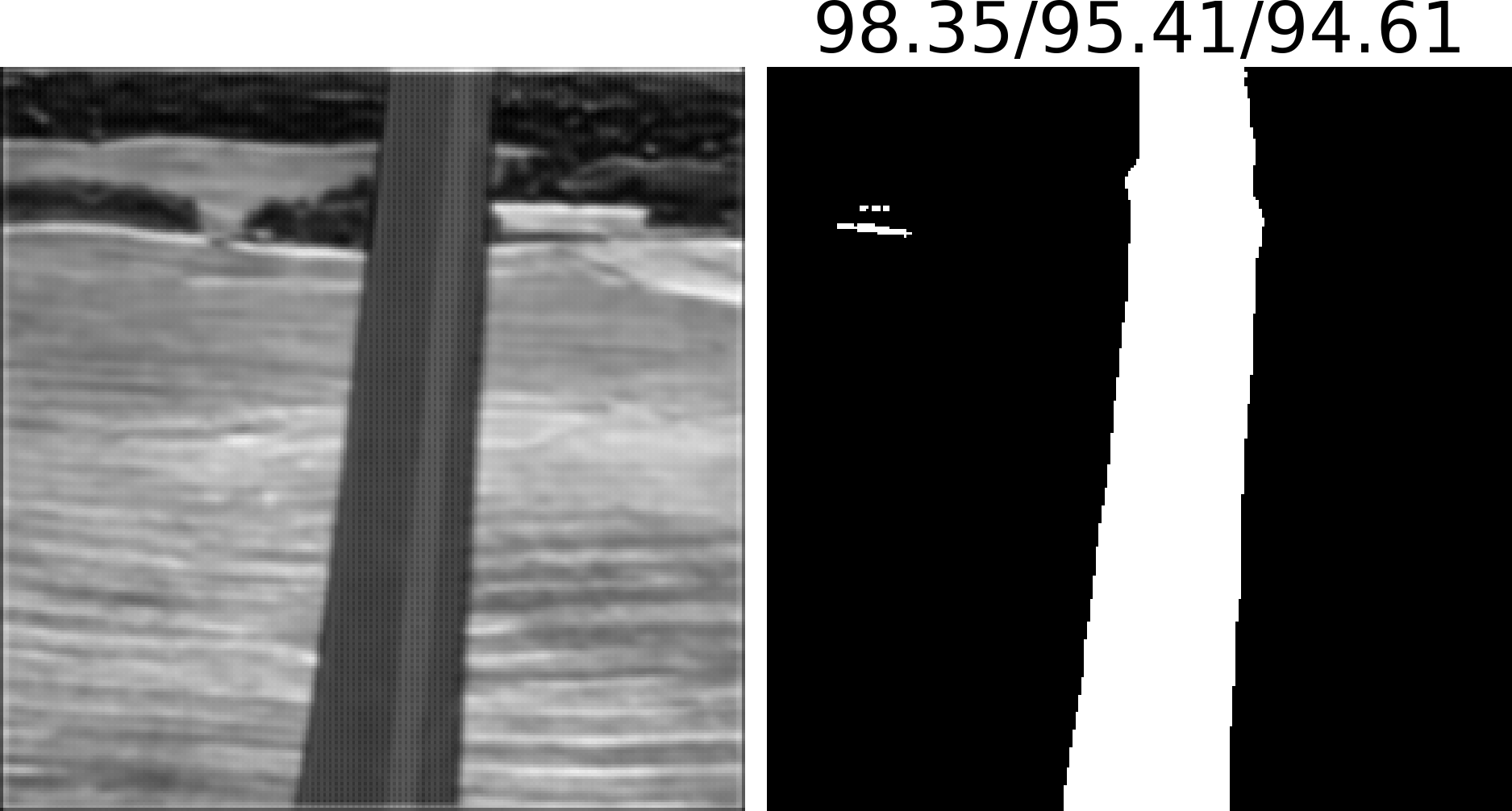}
        \end{subfigure} &
        \rotatebox{90}{\hspace{.38cm}\normalsize $d_{CS} = 4$}
        \begin{subfigure}[b]{0.71\linewidth}
            \centering
            \includegraphics[width=\linewidth]{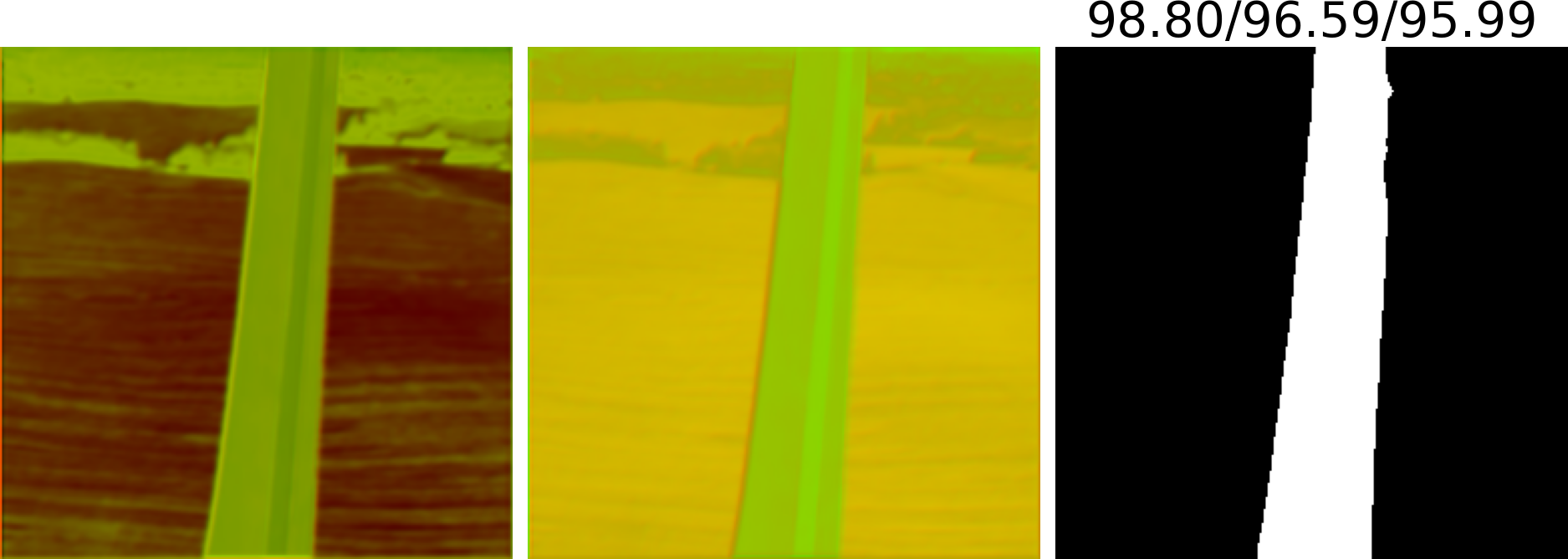}
        \end{subfigure} \\

        \rotatebox{90}{\hspace{.38cm}\normalsize $d_{CS} = 2$}
        \begin{subfigure}[b]{0.471\linewidth}
            \centering
            \includegraphics[width=\linewidth]{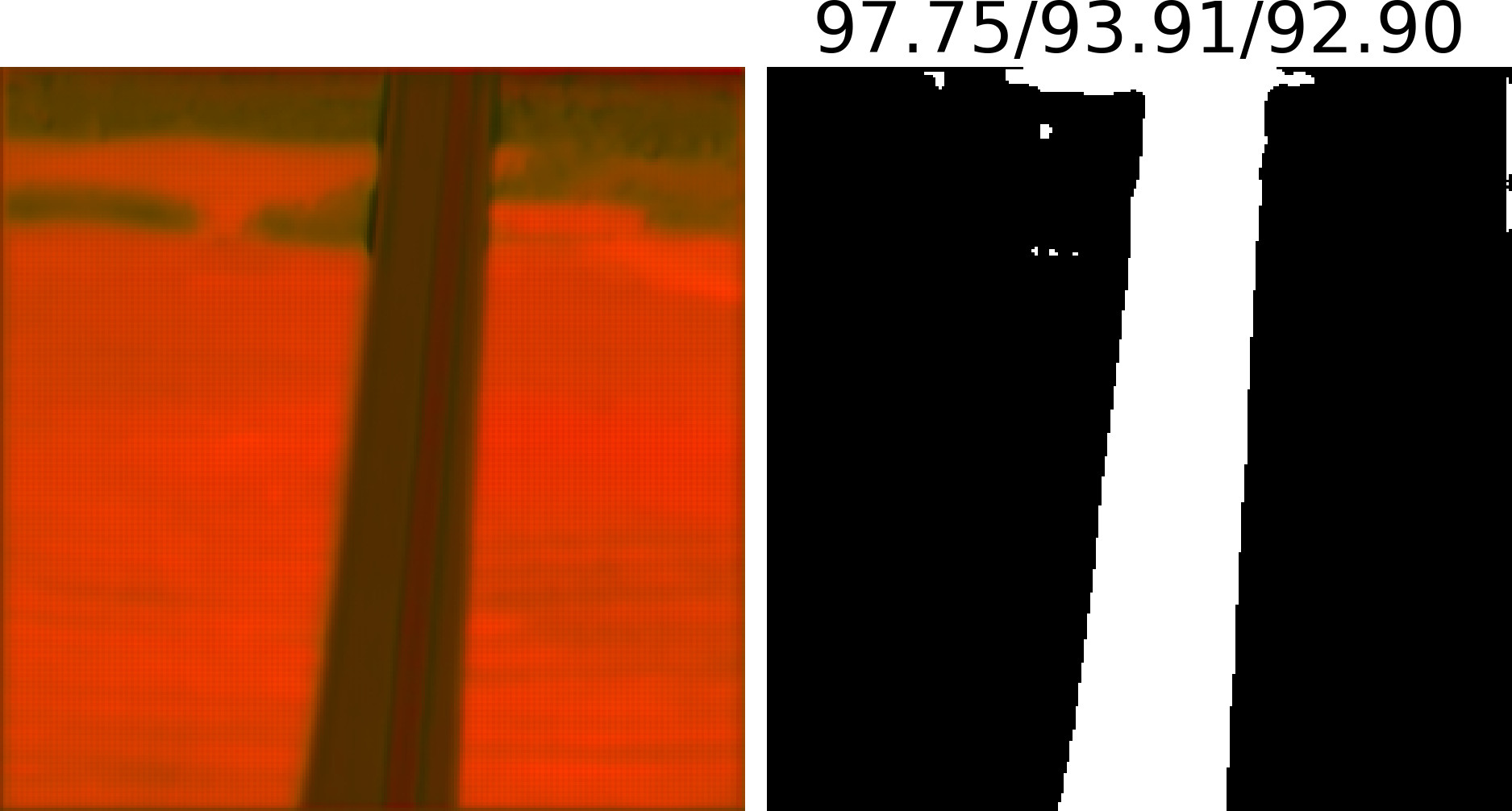}
        \end{subfigure} &
        \rotatebox{90}{\hspace{.38cm}\normalsize $d_{CS} = 5$}
        \begin{subfigure}[b]{0.71\linewidth}
            \centering
            \includegraphics[width=\linewidth]{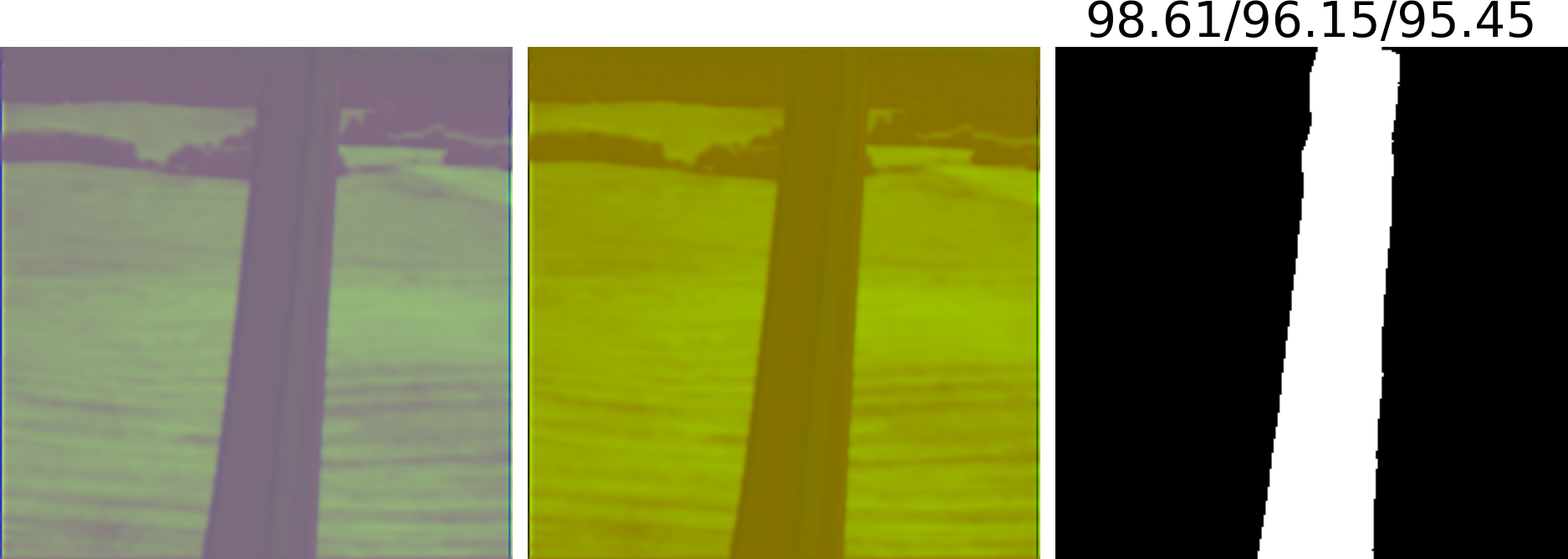}
        \end{subfigure} \\

        \rotatebox{90}{\hspace{.38cm}\normalsize $d_{CS} = 3$}
        \begin{subfigure}[b]{0.471\linewidth}
            \centering
            \includegraphics[width=\linewidth]{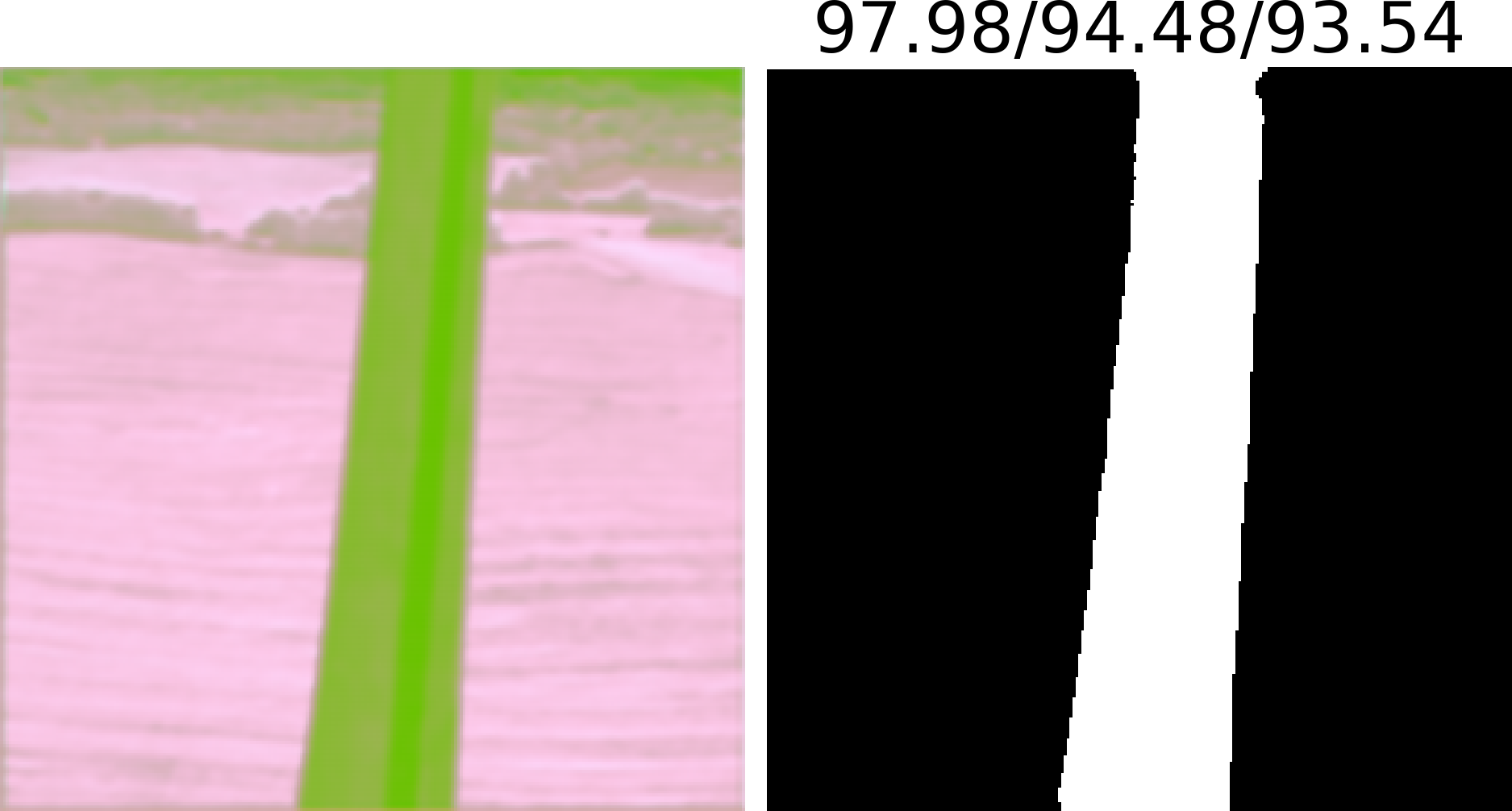}
        \end{subfigure} &
        \rotatebox{90}{\hspace{.38cm}\normalsize $d_{CS} = 6$}
        \begin{subfigure}[b]{0.71\linewidth}
            \centering
            \includegraphics[width=\linewidth]{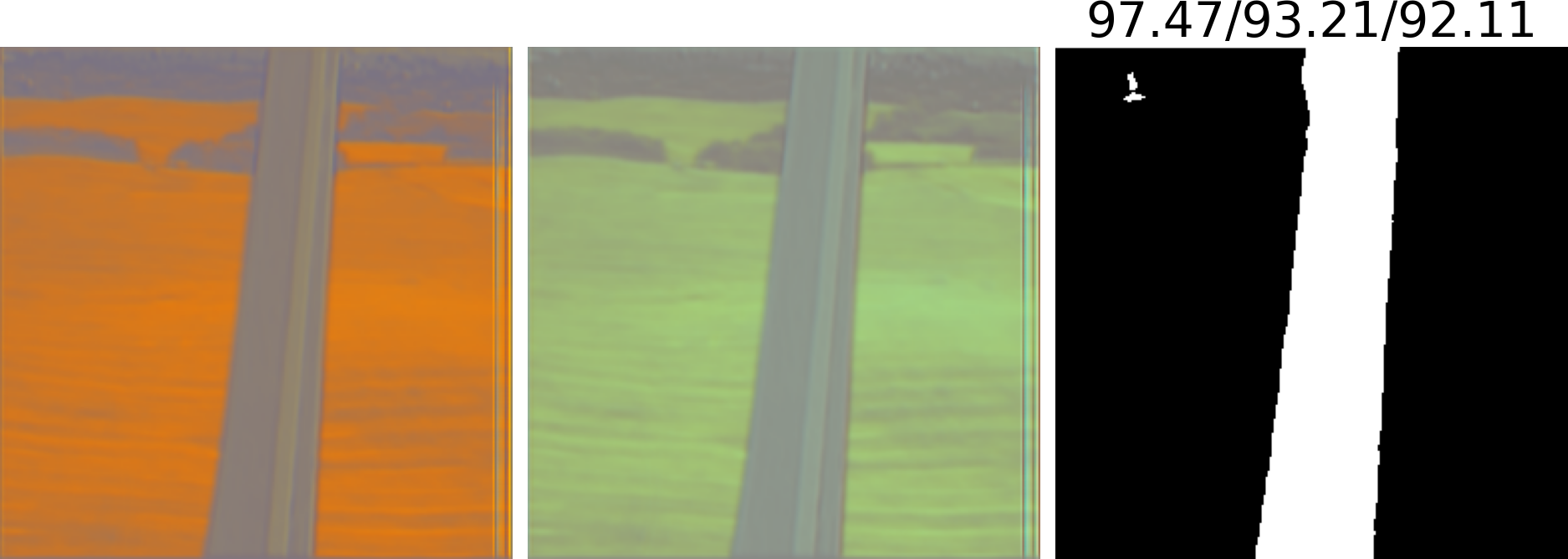}
        \end{subfigure} \\
    \end{tabular}
    \end{tabular}
    }
    \vspace{-0.4cm}
    \caption{\textbf{Visualization of colorspace transformations}. On the left side, RGB input (top) and ground-truth mask (bottom). On the right, the transformed image at various $d_{CS}$ for CSDA$^{(\Delta)}$ with its corresponding estimated mask on its right; accuracy, F1, and mIoU on top. See \cref{sec:visual_colorspace} for details.} 
    \label{fig:visual_colorspace}
    \vspace{-0.2cm}
\end{figure}

\begin{figure}[t!]
    \centering

    \begin{tabular}{@{}c@{}c@{}c@{}}
    \begin{subfigure}[b]{0.33\linewidth}
            \centering
            \includegraphics[width=\linewidth]{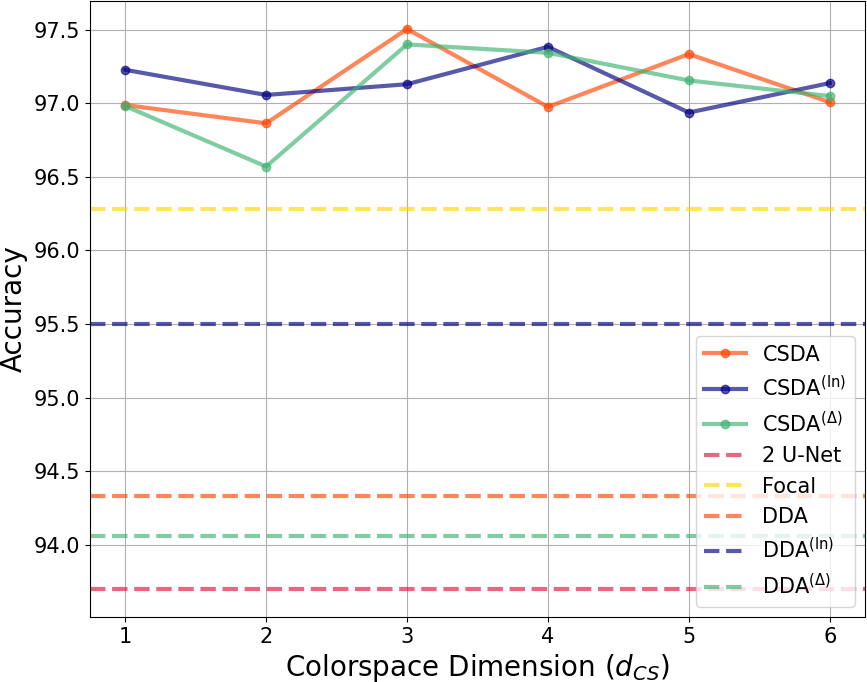} 
        \end{subfigure} &
        \begin{subfigure}[b]{0.33\linewidth}
            \centering
            \includegraphics[width=\linewidth]{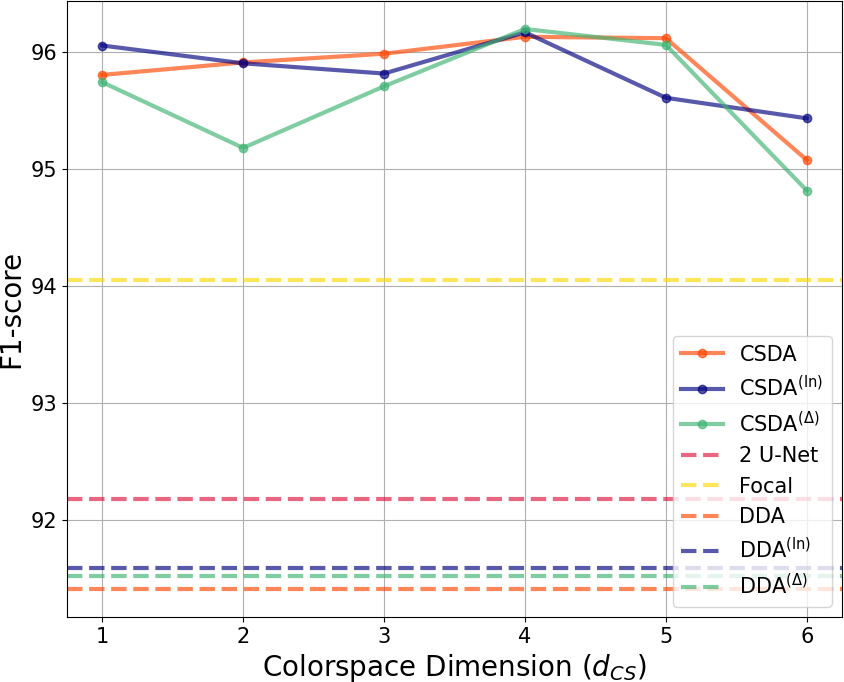}
        \end{subfigure}  &
        \begin{subfigure}[b]{0.33\linewidth}
            \centering
            \includegraphics[width=\linewidth]{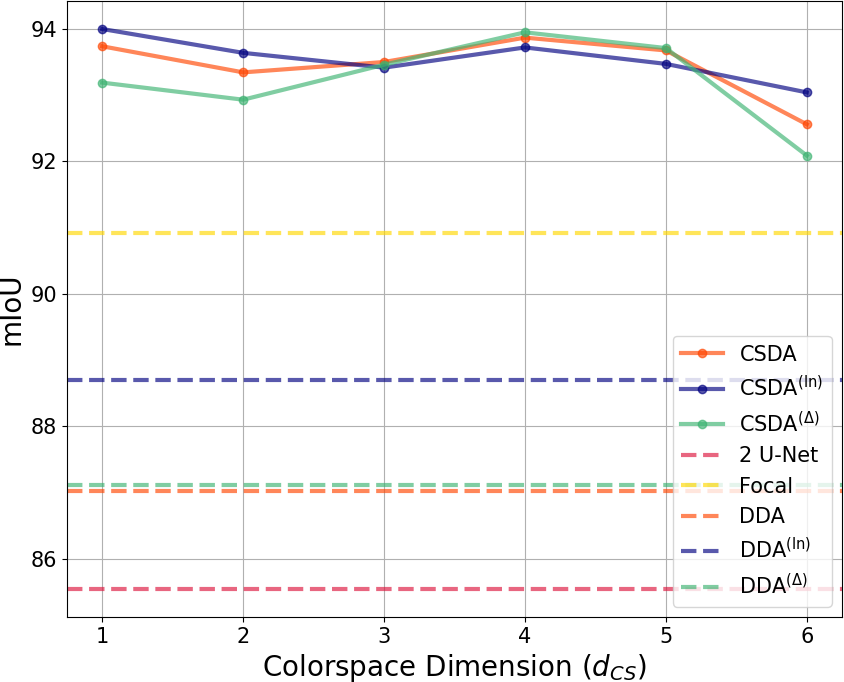}
        \end{subfigure}         
    
    \end{tabular}
        \vspace{-0.3cm} 
        \caption{\textbf{Segmentation performance across model variants and colorspace dimensions.} Detailed descriptions of the model variants are provided in \cref{sec:ablation}.} \label{fig:ablation}
        \vspace{-0.5cm}
\end{figure}

\vspace{-0.1cm}
\section{Experimental Results} \label{sec:results}
\vspace{-0.1cm}

We assess CSDA via an analysis of the learned colorspaces, an ablation on its color preprocessing, and quantitative and qualitative evaluations across different windfarms. Our data is comprised of images captured during drone turbine inspection following~\cite{bunet}. To emulate evaluation on newly acquired images, the test set is sourced from different windfarms.

\vspace{-0.15cm}
\subsection{Visual Exploration of Colorspaces} \label{sec:visual_colorspace}
\vspace{-0.15cm}

\Cref{fig:visual_colorspace} illustrates the colorspace transformation across dimensions $d_{CS}$ for CSDA$^{(\Delta)}$. For $d_{CS}=1$, the transformed image $\mathbf{Y}$ is shown in grayscale; for $d_{CS}=2$, as a partial RGB image with a zero-filled channel; and for $d_{CS}=3$, as a full RGB image. Higher dimensions are visualized by grouping channels into RGB triplets and padding with zeros if needed: $d_{CS}=4$ yields two partial RGB images, $d_{CS}=5$ one full and one partial RGB, and $d_{CS}=6$ two full RGB images.

At low colorspace dimensions, CSDA enhances contrast by brightening either the blade or background (e.g., at $d_{CS}=2$, the landscape is bright while the blade is dark). Increasing dimensionality enriches the color palette: at $d_{CS}=3$, blades appear green against a pink background; at $d_{CS}=6$, the background spans orange to green while blades remain gray. The model also reduces within-class variance, producing more uniform blade regions (e.g., the blade becomes consistently green at $d_{CS}=3$). Overall, CSDA adapts images into discriminative palettes where blade and background intensities are more clearly separated.
 
\vspace{-0.35cm}
\subsection{Ablation Study} \label{sec:ablation}
\vspace{-0.1cm}

\begin{figure}[t!]
\hspace{-0.2cm} \resizebox{1.03\linewidth}{!} {
\begin{tabular}{@{}c@{}}
\includegraphics[width=0.4\linewidth]{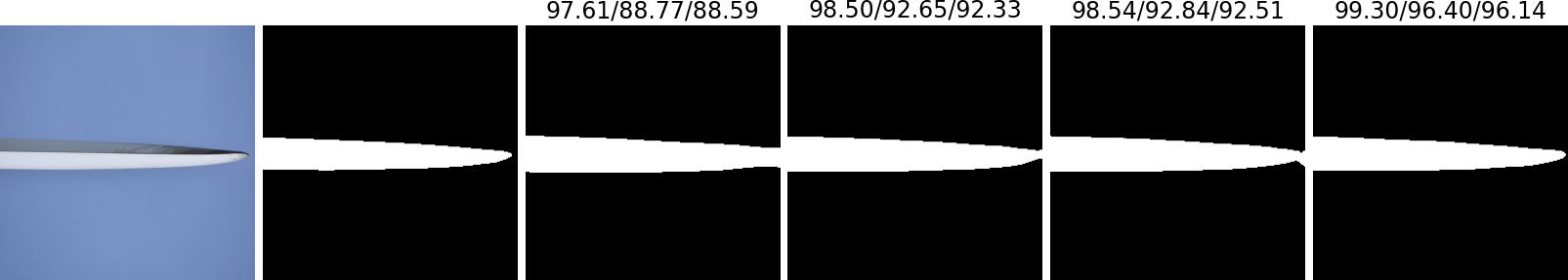} \vspace{-0.1cm} \\ 
\includegraphics[width=0.4\linewidth]{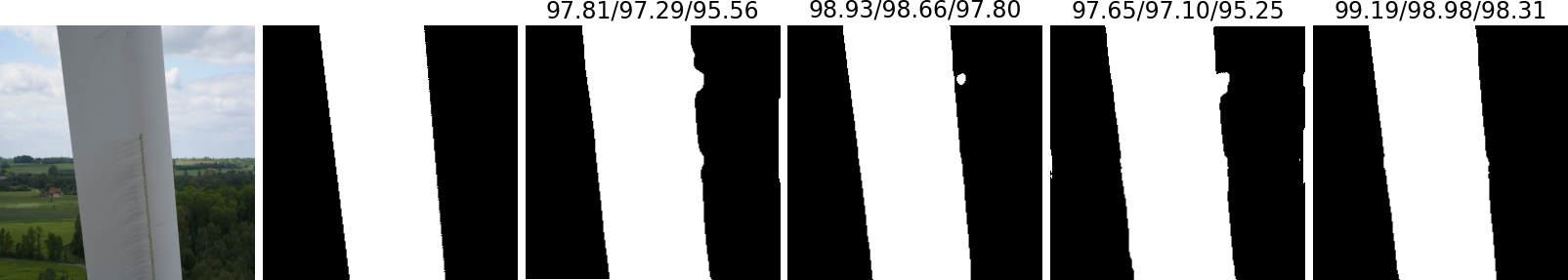} \vspace{-0.1cm} \\  
\includegraphics[width=0.4\linewidth]{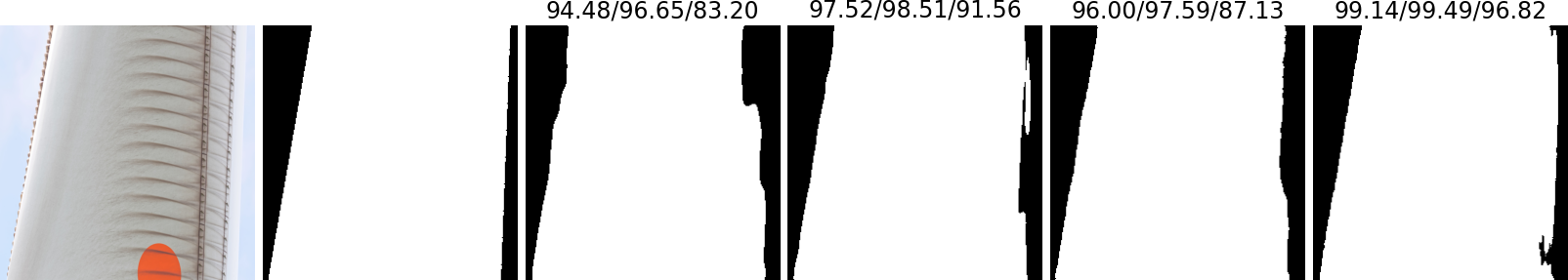} \vspace{-0.1cm} \\  
\includegraphics[width=0.4\linewidth]{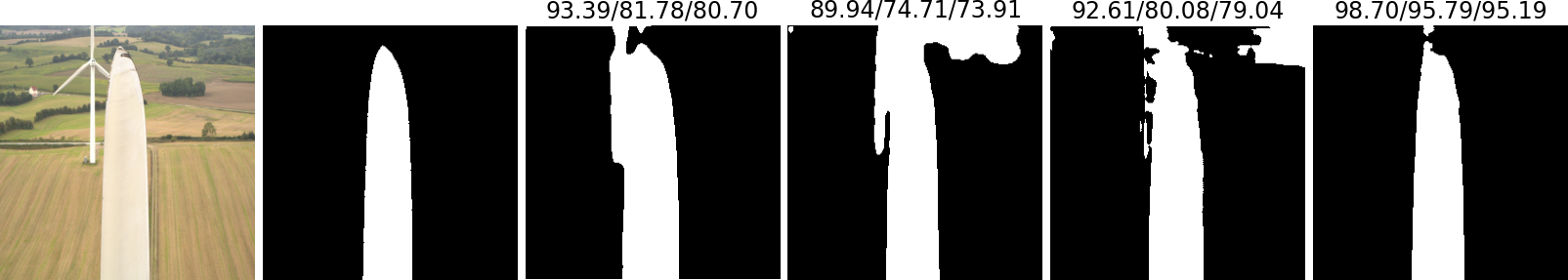} \vspace{-0.1cm}
\end{tabular}}
\vspace{-0.4cm}
\caption{\textbf{Qualitative comparison on test images for DDA, Focal and CSDA.} First two columns: input image and ground-truth mask $\mathbf{M}$. Next: estimated segmentation masks from DDA$^{(\Delta)}$, DDA$^{(\ln)}$, Focal, and CSDA$^{(\Delta)}$ with accuracy, F1, and mIoU shown above.}
\label{fig:visual_ablation}
\vspace{-0.6cm}
\end{figure}

We show that applying discriminant analysis for colorspace preprocessing before segmentation improves accuracy over using either discriminant analysis or segmentation alone. We compare against three baselines: (1) Deep Discriminant Analysis (DDA): maps RGB input to a 1D feature map ($d_{CS}=1$) using only the discriminant loss ($\mathcal{L}_{T} = \mathcal{L}_{DDA}$), with sigmoid outputs thresholded for validation accuracy; (2) Focal Model: a standard segmentation model trained directly on RGB images with focal loss ($\mathcal{L}_{T} = \mathcal{L}_{P}$); and (3) Two U-Net Model: CSDA's architecture but trained only with focal loss.

\Cref{fig:ablation,fig:visual_ablation} exhibit that combining discriminant analysis with segmentation consistently boosts performance in the blade domain. Quantitatively, as shown in \cref{fig:ablation}, CSDA outperforms both DDA and Focal baselines across all metrics, independent of the chosen loss. The gain is not architectural, as stacking U-Nets performs worst, but stems from effective discriminant preprocessing. Qualitative results (\cref{fig:visual_ablation}) confirm that CSDA better preserves blade contours and reduces background misclassifications. Finally, \Cref{fig:ablation} shows CSDA is robust to the choice of $d_{CS}$; we select $d_{CS}=4$ for subsequent experiments due to its consistently strong F1 and mIoU.



\begin{table}[t!]
\centering
\resizebox{\linewidth}{!} {
\begin{tabular}{lccccccccc}
\toprule
     \multicolumn{1}{c}{Method} & \multicolumn{1}{c}{Accuracy}  & \multicolumn{1}{c}{Precision} & \multicolumn{1}{c}{Recall} & \multicolumn{1}{c}{F1} & \multicolumn{1}{c}{mIoU} & \multicolumn{1}{c}{IoU$_{C_0}$} & \multicolumn{1}{c}{IoU$_{C_1}$} \\
    \multicolumn{1}{c}{} & {[\%]} & {[\%]} & {[\%]}  & {[\%]}  & {[\%]} & {[\%]}  & {[\%]}  \\
    \midrule
    DeepLabv3+~\cite{deeplabv3+}  & 94.14 & 96.36 & 87.38 & 89.03 & 87.47 & 90.31 & 84.62 \\  
    SW~\cite{sw} & 93.48 & 93.57 & 91.71 & 91.37 & 87.44 & 88.64 & 86.23 \\ 
    ResNeSt~\cite{resnest} & 94.23 & 96.84 &91.47 & 92.77 & 89.63 & 90.40 & 88.86 \\ 
    U-NetFormer~\cite{unetformer} & 96.20 & 97.31 & 93.51 & 94.42 & 91.75 & 92.53 & 90.96 \\ 
    SAM~\cite{sam} & 94.36 & 97.29 & 91.22 & 92.60 & 91.66 & 92.31 & 91.01  \\ 
    CLIPSeg~\cite{clipseg} &  82.70 & 77.02 & 75.52 & 74.29 & 75.09 & 80.16 & 70.02 \\
    DiffSeg~\cite{diffseg} & 96.37 & 82.08 & 89.74 & 85.73 & 86.40 & 91.66 & 81.13 \\
    EfficientFormer~\cite{efficientformer} & 96.42 & 95.47 &  93.63 & 94.55 & 93.51 & 94.02 & 92.99 \\ 
    MobileViT~\cite{mobilevit} & 96.14 & 95.44 & 93.33 & 94.38 & 93.47 & 94.06 & 92.88\\
    Mask2Former~\cite{mask2former} & 96.68 & 95.63 & 93.89 & 94.76 & 93.72 & 94.49 & 92.93\\
    BU-Net~\cite{bunet} & \underline{97.39} & \textbf{99.42} & 93.35 & 95.73 & 93.80 & \textbf{94.70} & 92.90\\ 

   CSDA & 96.98 & \underline{96.78} & 96.65 & 96.13 & \underline{93.86} & 94.21 & \textbf{93.51}   \\ 
   CSDA$^{(\ln)}$ & \textbf{97.40} & 95.72 & \textbf{97.38} & \underline{96.17} & 93.72 & 94.25 & 93.18\\
   CSDA$^{(\Delta)}$ & 97.34 & 95.97 & \underline{97.30} & \textbf{96.20} & \textbf{93.94} & \underline{94.51} & \underline{93.38} \\ 

\bottomrule
\end{tabular}
}
\vspace{-0.3cm}
\caption{\textbf{Quantitative comparison} on blade segmentation.} \label{tab:seg-unet-compare}
\vspace{-0.5cm}
\end{table}

\vspace{-0.15cm}
\subsection{Quantitative Evaluation}
\vspace{-0.1cm}
\Cref{tab:seg-unet-compare} compares CSDA models from \cref{sec:losses} against competing segmentation models. CSDA$^{(\ln)}$ and CSDA$^{(\Delta)}$ show slight improvements over the standard CSDA, benefiting from adjustable class variances via $\lambda_F$. Moreover, they achieve top-tier performance, maintaining balanced precision and recall. CSDA$^{(\ln)}$ matches the highest accuracy alongside BU-Net~\cite{bunet}, and CSDA$^{(\Delta)}$ achieves the best F1-score and mIoU-metrics that offer robust and consistent evaluation across hyperparameters like $d_{CS}$, as discussed in \cref{sec:ablation}.

Computationally, CSDA remains efficient at inference time, as the discriminant losses are only computed during training. While CSDA stacks two U-Nets, each U-Net is lightweight, with a single model requiring just 0.024s per image on an NVIDIA RTX 3080 Ti and Intel i9. Thus, CSDA’s inference time requires only 0.048s, significantly faster than BU-Net~\cite{bunet}, the next-best performer, which takes 0.089s.


\vspace{-0.15cm}
\subsection{Qualitative Analysis} \label{sec:qualitative}
\vspace{-0.15cm}

\begin{figure}[t!]
\hspace{-0.1cm}\resizebox{1.02\linewidth}{!} {
 \begin{tabular}{@{}c@{\hspace{0.13cm}}:c@{\hspace{0.13cm}}@{}}

\includegraphics[width=0.55\linewidth]{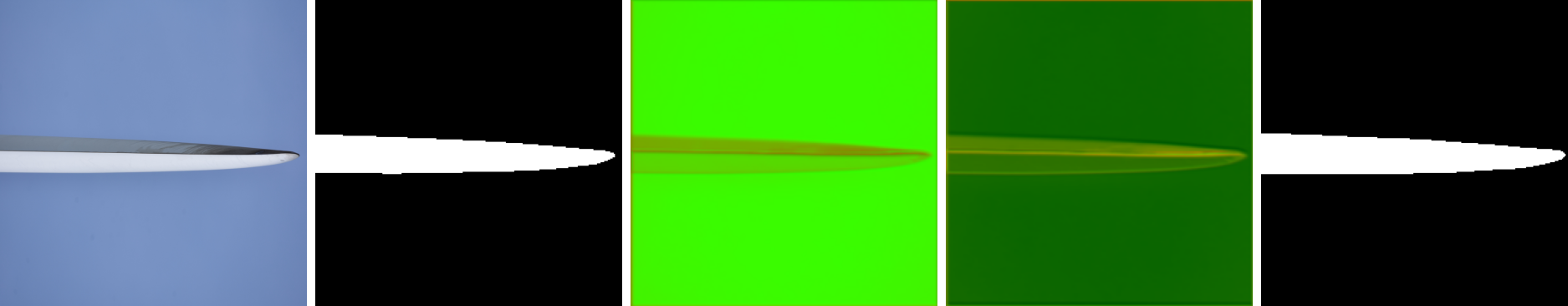}&
\includegraphics[width=0.44\linewidth]{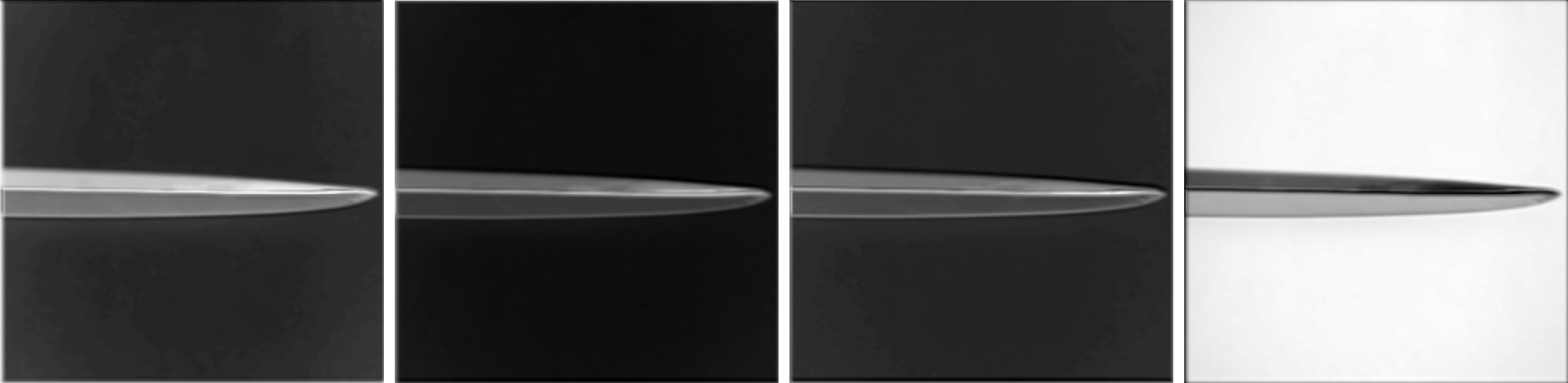}\\

\includegraphics[width=0.55\linewidth]{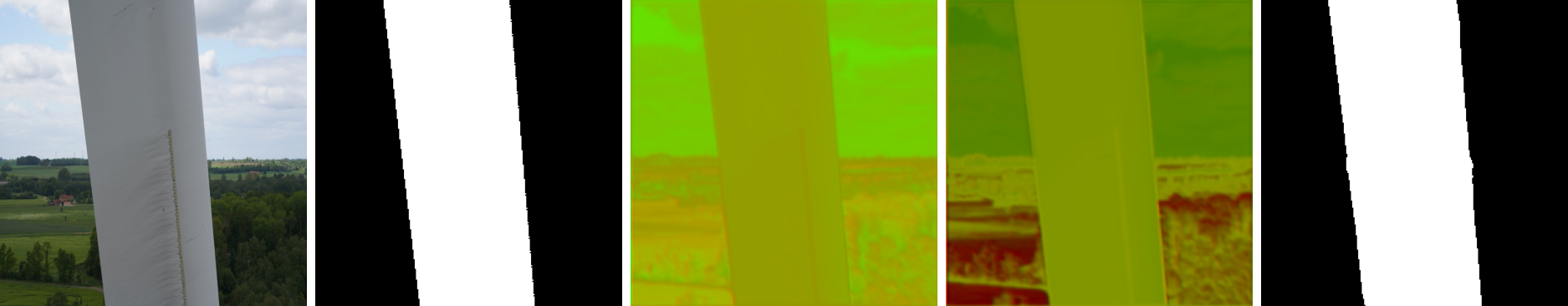}&
\includegraphics[width=0.44\linewidth]{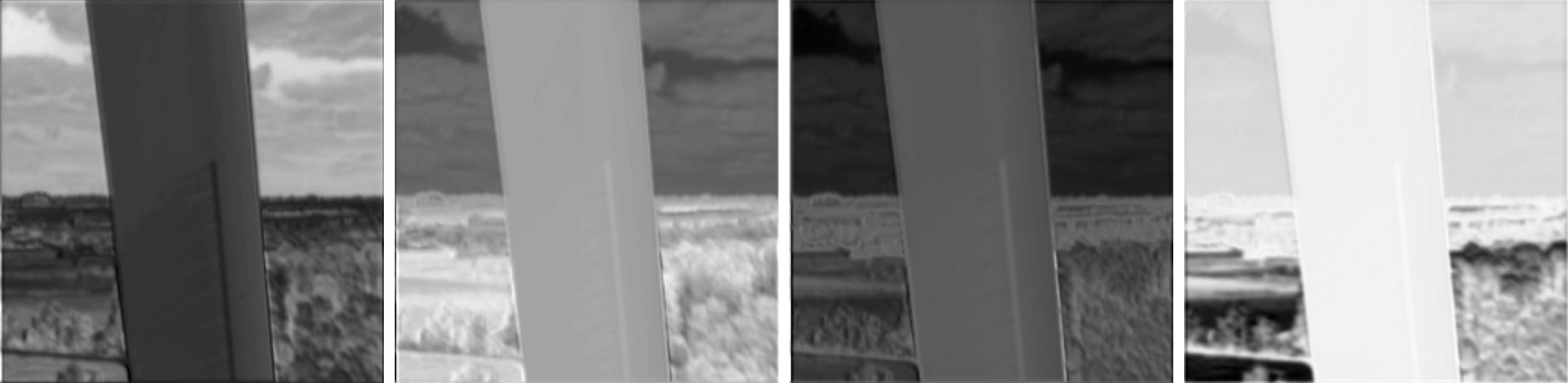} \\

\includegraphics[width=0.55\linewidth]{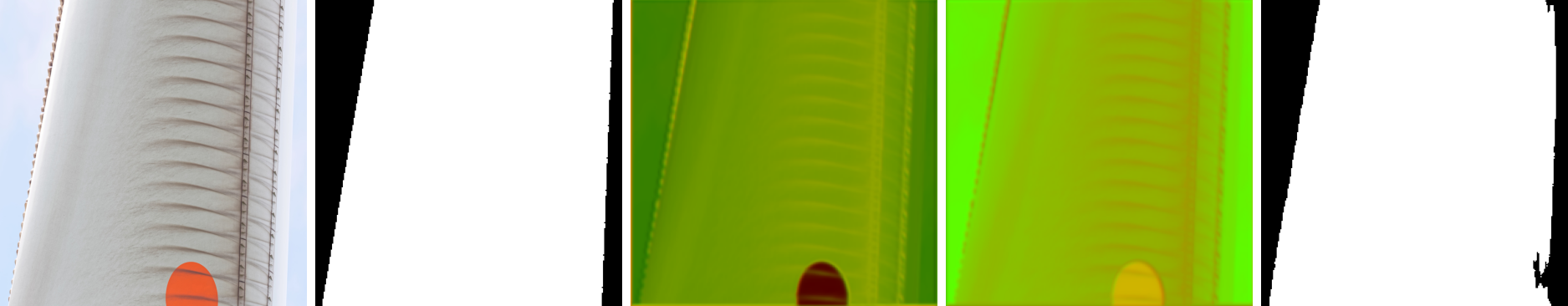}&
\includegraphics[width=0.44\linewidth]{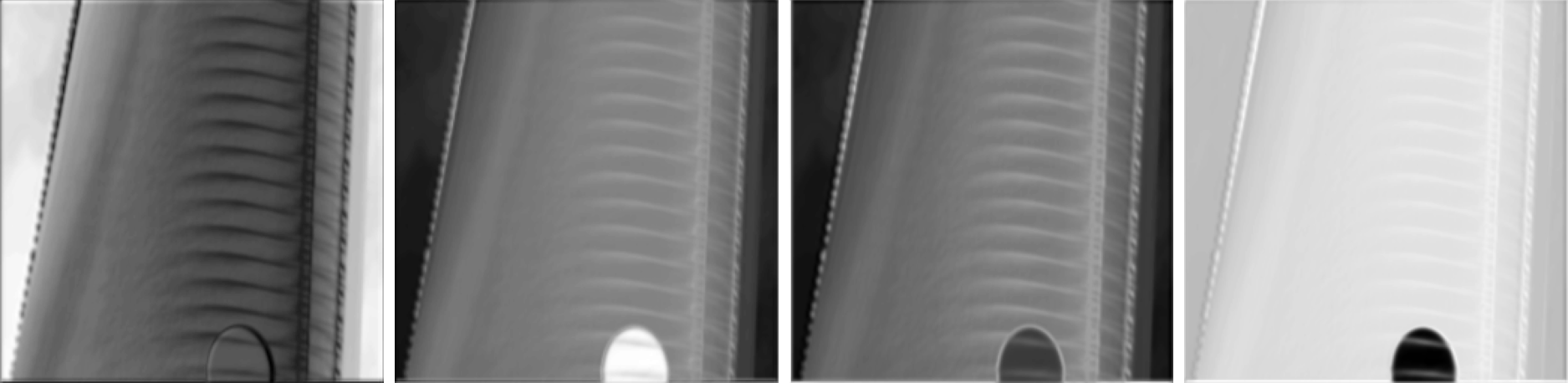}\\

\includegraphics[width=0.55\linewidth]{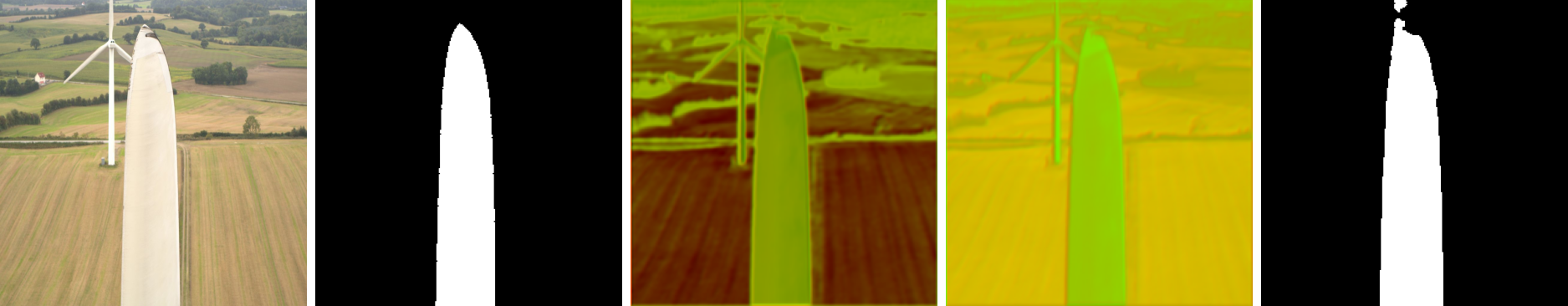}&
\includegraphics[width=0.44\linewidth]{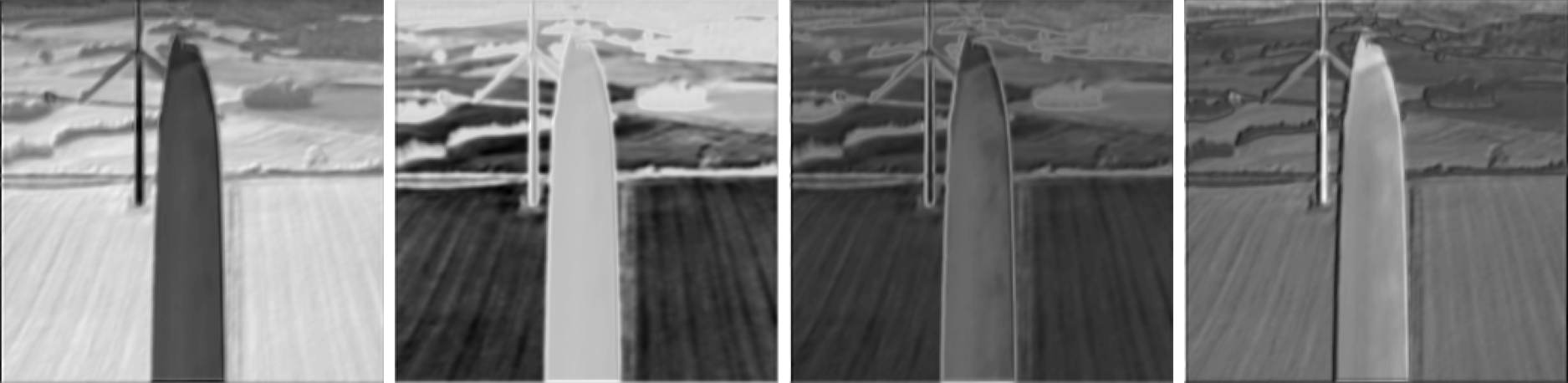}
\end{tabular}}
\vspace{-0.44cm}
\caption{\textbf{Qualitative results for CSDA$^{(\Delta)}$ with $d_{CS}=4$.} First two columns: input image and ground-truth mask $\mathbf{M}$. Third to fourth column: transformed colorspace image, see \cref{sec:visual_colorspace}. Fifth column: estimated mask. Last columns: individual transformed colorspace channels.} \label{fig:qualitative}
\vspace{-0.3cm}
\end{figure}

We illustrate the effectiveness of CSDA$^{(\Delta)}$ with $d_{CS}=4$ in enhancing segmentation. \Cref{fig:qualitative} shows the RGB input and ground-truth mask, followed by the colorspace transformation $\mathbf{Y}$ and the predicted segmentation. $\mathbf{Y}$ is visualized both as partial RGB images (using two colorspace channels plus a zero channel) and as individual grayscale channels.

The discriminative power of the CSDA transformation is evident: the blade area becomes more homogeneous due to within-class variance is reduced, while between-class variance increases, helping separate the blade from the background. For example, in the first channel of the third instance, CSDA effectively maps the orange spot on the blade to similar intensities as the rest of the blade, aiding segmentation. In more complex scenes, such as the last example where the blade and a background turbine share similar RGB colors, CSDA still isolates the foreground blade accurately, capturing the blade tip, something BU-Net fails to achieve~\cite{bunet}.

\vspace{-0.1cm}
\subsection{Generalization across Windfarms}
\vspace{-0.1cm}

\begin{figure}[t!]
    \centering
        \includegraphics[width=0.87\linewidth]{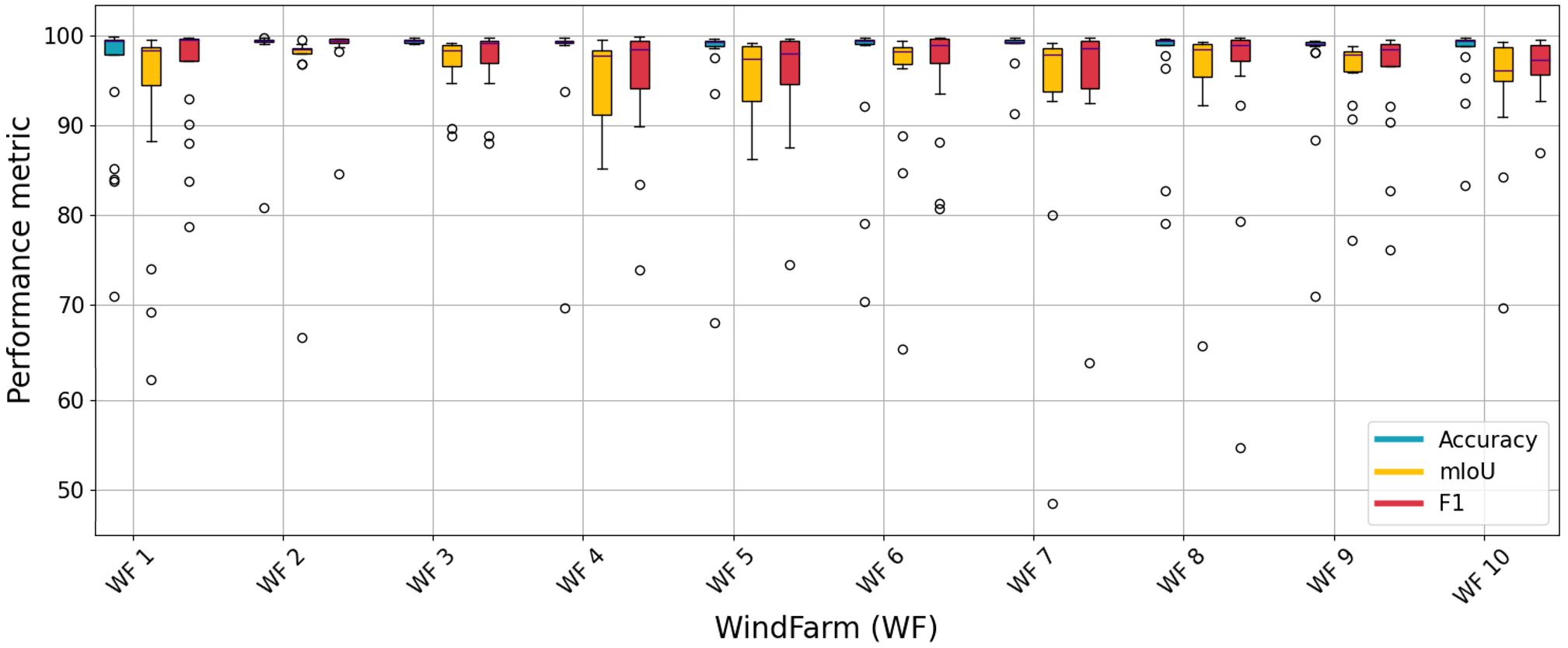} 
    \vspace{-0.4cm} \caption{ \textbf{Boxplot of the segmentation results} across diverse windfarms of the test set for CSDA$^{(\Delta)}$ with $d_{CS}=4$.} \label{fig:windfarm}
    \vspace{-0.6cm}
\end{figure}

To assess the robustness of CSDA$^{(\Delta)}$ with $d_{CS}=4$, we evaluate its performance across multiple windfarm inspections included exclusively in the test set. \Cref{fig:windfarm} illustrates a consistently high performance across all windfarms, indicating strong generalization to images in our real-world setting. 

    

\vspace{-0.25cm}
\section{Conclusion}
\vspace{-0.15cm}

This paper presents CSDA, a novel colorspace transformation optimized for image segmentation preprocessing. CSDA extends LDA by jointly optimizing a multidimensional nonlinear colorspace transformation with the segmentation objective, emphasizing discriminative color features relevant to the target domain. To our knowledge, CSDA is the first approach to integrate nonlinear discriminant analysis into a learnable colorspace transformation for segmentation. Experimental results on a real industry problem demonstrate the effectiveness of colorspace optimization in domain-specific segmentation. Significant improvements in wind turbine blade segmentation are observed compared to optimizing either segmentation loss or discriminant analysis alone. These results underscore the value of task-specific preprocessing and establish CSDA as a robust tool for domain-adapted segmentation tasks.



\bibliographystyle{IEEEbib}
\bibliography{strings}

\end{document}